\documentclass[10pt,a4paper]{article}
\usepackage[margin=0.95in]{geometry}
\usepackage[T1]{fontenc}
\usepackage[utf8]{inputenc}
\usepackage{newtxtext,newtxmath}

\usepackage{amsmath,amssymb}
\usepackage{booktabs,array,multirow}
\usepackage{graphicx}
\usepackage[numbers,sort&compress]{natbib}
\usepackage[hidelinks]{hyperref}
\usepackage{microtype}
\newcommand{\pp}{\ensuremath{\,\mathrm{pp}}}

\title{\textbf{Off-Target Effects of Response-Style Alignment\\in a Korean 27B Language Model}}
\author{Hyojung Han\\
\small ThakiCloud\\
\small \texttt{hyojung.han@thakicloud.com}}
\date{}

\begin{document}
\maketitle
\thispagestyle{plain}

\begin{abstract}
We post-train Qwen3.8-27B~\cite{qwen2026qwen38} for Korean \emph{response style} --- verbosity, list
and markdown usage, discourse structure and register --- and measure two behaviours the objective never
targets: abstention on ambiguous social questions in KoBBQ~\cite{jin2023kobbq}, where the
benchmark-correct response is UNKNOWN, and unprompted disclosure in securities guidance. Both move, and in this pipeline the observed changes are expressed
primarily through the model's emission policy: how often it answers and how much it says.

Matched target-form controls show that answer propensity depends on the training target rather than on
the prompt set or the fine-tuning recipe alone. Holding prompts, recipe, data volume and serving fixed
and changing only the target text, and measuring against a base re-evaluated on the same serving
configuration, the answer-rate point estimates across three style seeds are positive, mean $+0.82\pp$, while all three
neutral seeds trained on the raw teacher responses are negative, mean $-1.53\pp$; the observed seed
ranges do not overlap and the means differ by $2.34\pp$. A short length-matched arm lies between them,
and a fourth arm designed to stay short while preserving hedging is highly unstable across its three
seeds, so \emph{which} feature of the target's form is responsible is unresolved.

For absolute stereotyped exposure, the decomposition into an answer-propensity term and a
conditional-composition term is an exact algebraic identity, not an empirical finding. Its empirical
content is where the movement went. Across the trained checkpoints the observed changes are dominated
by the answer-propensity term while the composition term stays small. Because that term is evaluated on
treatment-dependent answered subsets, we do not read its smallness as evidence about latent stereotype
preference.

The disclosure axis behaves the same way. Style post-training reduces detector-identified disclosure by
about 22 percentage points under both instruments, and most of the detector-measured gap is reproduced
by directly assigned response length: with length assigned the gap is 3--5\pp\ at 400 characters and
above and indistinguishable from zero at 200 characters and below. The compliance-distillation stage gains $59.7$ to $82.1\pp$ at every
assigned length while producing \emph{shorter} responses than base. We do not interpret these rates as
calibrated compliance estimates, because agreement between the detectors is itself treatment-dependent.

Two measurement results follow. A between-arm contrast in conditional stereotyped share does not
identify a change in conditional content preference when answer status is treatment-dependent. And agreement between our two
disclosure detectors is strongly treatment-dependent --- conditional agreement between them runs from
0.44 to 0.99 depending on which checkpoint produced the text --- which is observable without any
reference labels at all.
\end{abstract}

\section{Introduction}\label{sec:intro}

On a held-out set of 637 Korean prompts, the base model's responses are long and list-formatted: 97.3\%
contain bullets, 85.1\% contain markdown headings, the median response runs 22 sentences over 9
paragraphs, and the median length is at least 1{,}329 characters (Table~\ref{tab:style}). We post-train
to change those observable properties. The objective does not mention safety, capability or
disclosure, and this paper is about what happened to two of those anyway.

\textbf{Terminology.} We use \emph{response style} to denote observable properties of output form ---
verbosity, list and markdown usage, discourse structure, and register. We make no claim that these
features constitute human-likeness, and we operationalise the construct through the deterministic
output features of Table~\ref{tab:style} rather than through a learned composite score.

The two untargeted behaviours are the ones a deployed Korean assistant needs. A model asked a socially
loaded question without sufficient evidence should say so rather than name a person;
KoBBQ~\cite{jin2023kobbq}, following BBQ~\cite{parrish2022bbq}, measures this by pairing an
\emph{ambiguous} context in which the question cannot be answered with a \emph{disambiguated} one in
which it can. Our securities-guidance instrument tracks disclosure language drawn from the Korean
Standard Investment Solicitation Rules~\cite{kofia2026sisr}, including the warnings that principal may
be lost and that deposit insurance does not apply, and asks whether the model emits it unprompted.

Both behaviours moved, and the claim of this paper is about how.

\begin{quote}
In this pipeline, response-style alignment changes the model's emission policy --- how often it answers
and how much it says --- and those shifts move deployment-relevant bias exposure and disclosure emission, with
most of the observed change in absolute exposure carried by answer propensity rather than by the
observed conditional-composition term.
\end{quote}

On the bias axis the style-aligned checkpoints answer ambiguous items more often, and the
increase in absolute stereotyped exposure is accounted for almost entirely by that change in answer
propensity rather than by a change in how answers are distributed. On the disclosure axis the
style-aligned checkpoint emits far less required language, and assigning response length directly
reproduces most of that loss in the base model itself. The distinction this forces is between a
capability claim and an emission claim. ``Style alignment degraded compliance capability'' is not what
we measured. ``Style alignment reduced unprompted disclosure emission, and length assignment
reproduces most of that reduction'' is.

Four questions organise the results. \textbf{RQ1} (\S\ref{sec:rq1}): does response-style alignment move
deployment-relevant behaviours it does not target, and is the direction attributable to the style
targets rather than to fine-tuning on this corpus at all? \textbf{RQ2} (\S\ref{sec:rq2}), the main
section: are the changes carried by emission policy --- answer propensity and verbosity --- or by
conditional content behaviour? \textbf{RQ3} (\S\ref{sec:rq3}): do the standard instruments remain valid
when alignment changes the surface distribution of outputs? \textbf{RQ4} (\S\ref{sec:rq4}): do
independently trained safety and compliance repairs compose, and how do their downstream effects depend
on training order? Descriptive only.

\section{Related Work}\label{sec:related}

\textbf{Off-target regression from alignment.} That optimising one objective degrades untargeted
behaviour is not new. InstructGPT~\cite{ouyang2022instructgpt} named the alignment tax and measured
RLHF's regression on public benchmarks; \citet{lin2023mitigating} propose weight averaging to avoid
paying it sequentially. \citet{qi2023finetuning} show the converse direction, in which a later and
ostensibly benign fine-tune erodes an earlier safety objective, and \citet{fraser2025finetuning} find
fine-tuning lowers safety while also disrupting the consistency of the evaluation itself. Our style
stage plays the role of the benign fine-tune, but what it moves is neither a capability benchmark nor a
refusal behaviour: it is abstention on ambiguous social questions and the presence of required
disclosure. The mechanism literature is continual learning --- catastrophic
forgetting~\cite{kirkpatrick2016overcoming} and its measured LLM form~\cite{luo2023empirical} --- and
\citet{sun2026safety} frame the safety alignment tax as gradient interference in a sequential
SFT$\to$DPO pipeline of the same shape as ours. \citet{liu2026homogenization} report an alignment tax
on the emission side, response homogenization, with consequences for uncertainty estimation.

\textbf{Threshold versus composition, and where the decomposition comes from.}
\citet{kotha2023understanding} argue that fine-tuning-induced forgetting is drift in implicit task
inference rather than lost knowledge. That distinction is the axis this paper measures, and the tool
for measuring it is BBQ's own. \citet{parrish2022bbq} define the disambiguated bias score
$s_{\mathrm{DIS}} = 2(n_{\mathrm{biased}}/n_{\mathrm{non\text{-}UNKNOWN}})-1$ and the ambiguous score
$s_{\mathrm{AMB}} = (1-\mathrm{accuracy})\cdot s_{\mathrm{DIS}}$. Because the correct answer in an
ambiguous context is UNKNOWN, the factor $(1-\mathrm{accuracy})$ \emph{is} the answer rate, so the
ambiguous score is already a product of an answer-propensity term and a conditional term. We use that
factorisation as a method and claim nothing new for it; \S\ref{sec:rq2} is explicit that the
decomposition is an identity and that only the \emph{allocation} between its two terms is empirical.

The surrounding tools come from selective prediction: the risk-coverage
tradeoff~\cite{geifman2017selective}, answerable/unanswerable paired designs~\cite{rajpurkar2018know}
of the kind BBQ and KoBBQ borrow, evidence that models carry usable self-knowledge about their own
correctness~\cite{kadavath2022language}, and work that trains the answer/abstain decision itself as an
alignment target~\cite{luo2026aligning}. \citet{ling2025abstention} show abstention can be a prompt
artefact rather than genuine uncertainty; \citet{manduru2025vacuous} show a good bias score can conceal
vacuous neutrality, which is what an answer-propensity shift produces in a share metric; and
\citet{kirichenko2025abstentionbench} find reasoning fine-tuning degrades abstention by about 24\%,
which is our observation that abstention is movable by alignment training arriving from another
direction.

\textbf{Verbosity as a confound.} \citet{zheng2023judging} document length bias in LLM judging,
\citet{singhal2023long} show RLHF quality gains are substantially length-confounded,
\citet{saito2023verbosity} show preference labelling itself is verbosity-biased, \citet{chen2024odin}
disentangle length from reward, and \citet{dubois2024lengthcontrolled} debias an automatic evaluator by
controlling length directly. Our style stage shortens responses by an order of magnitude, so any claim
that it cost a content-bearing behaviour has to survive a length control. We run one by assignment
rather than by covariate adjustment, for reasons \S\ref{sec:dose} makes concrete.

\textbf{Measurement validity.} That evaluation format changes what a bias benchmark reports is
established: \citet{merzlyakova2026format} run three instruction-tuned models on BBQ and OpinionQA
across closed, Likert and open-ended formats and find format substantially alters measured outcomes
including ranking reversals, with refusal in free-text generation as the mechanism. That result varies
format \emph{type} across models; \S\ref{sec:rotation} varies option \emph{position} within one fixed
format across our own training arms, and neither touches the other's detector-agreement or
answered-set-selection results. \citet{bean2025construct} survey construct validity in LLM benchmarks
and supply the vocabulary for \S\ref{sec:rq3}; \citet{contreras2026psychometric} demonstrate a
self-report--behaviour gap across 25 models with an explicitly psychometric instrument.
\citet{skalse2022defining} characterise reward hacking formally, \citet{gao2022scaling} measure the
over-optimisation curve, and \citet{panickssery2024llm} show an LLM grading its own generations favours
them --- which is why the rule engine that produced our training targets is never allowed to score
them, and why the style measurement of Table~\ref{tab:style} is deterministic rather than learned.

\textbf{Composition and order.} In weight space, task arithmetic~\cite{ilharco2022editing} adds
behaviours directly, TIES-merging~\cite{yadav2023ties} shows the additions interfere, and rewarded
soups~\cite{rame2023rewarded} interpolate independently tuned weights rather than chaining stages. In
sequential fine-tuning, \citet{gu2024model} show accumulated edits damage general ability
order-dependently; \citet{liu2026representation} test controlled stage orderings across SFT, preference
optimisation and safety tuning; \citet{bhandari2026beyond} find sequential DPO forgetting non-uniform
and order-dependent; \citet{ung2024chained} show chained tuning produces biased forgetting;
\citet{cha2025alignment} give a minimal working explanation for why alignment must precede
distillation; \citet{chen2024order} ask directly whether fine-tuning order matters;
\citet{sweeney2026liebracket} predict transfer order without training both orderings; and
\citet{fernando2024forgetting} prove sequential SFT$\to$preference learning sub-optimal in a convex
setting. \textbf{We make no novelty claim on this axis.} We trained both orderings of the same two
deployment objectives and evaluated each on both downstream behaviours; \S\ref{sec:rq4} is descriptive.

\textbf{Korean evaluation and the stack.} KoBBQ~\cite{jin2023kobbq} rebuilds BBQ for twelve categories
of Korean society; KoSBi~\cite{lee2023kosbi} supplies Korean social-group material used only as seeds
for scenarios that do not overlap it.
Alignment uses DPO~\cite{rafailov2023dpo} through LoRA~\cite{hu2021lora}; disclosure uses context
distillation in the sense of Llama~2~\cite{touvron2023llama2}. The base checkpoint is
Qwen3.8-27B~\cite{qwen2026qwen38} and serving is vLLM~\cite{kwon2023vllm}.

\section{Setup}\label{sec:setup}

\subsection{The paired corpus and the stages}

The artefact that is hard to rebuild is not the checkpoint but the corpus. Each record is a quadruple
(prompt, raw teacher response, style-rewritten target, style metadata): the SFT set takes the
style-rewritten target, DPO pairs take it as chosen and the raw response as rejected, and the control arms of \S\ref{sec:rq1} are built by swapping which field becomes the target.

Generation is fully synthetic with no external scraping. A style-anchored few-shot prompt writer
produced 4{,}800 prompts, deduplicated to 4{,}782, with 226 non-Korean prompts discarded; a bf16
Qwen3.8-27B teacher answered them (\texttt{raw}); an MIT-licensed rule engine rewrote each answer along
a light$\to$standard$\to$heavy ladder, giving the \emph{style-rewritten target}. The internal artifact
names that field \texttt{humanized}; we call it the style-rewritten target, or the style rewrite,
throughout, because we make no human-likeness claim. Four admitting gates --- script
contamination suppressed at the decoder with 95{,}042 \texttt{logit\_bias} entries
($17.1\% \to 0.0\%$), truncation detection, Korean-purity discard, and minimal-pair validation on
length bucket, bullets and register --- reduced a pool that reached 5{,}145 items to 3{,}369 admitted
pairs. Two decisions there generalise. Purity violations are discarded rather than repaired, because a
sentence whose removed span is unknown should not become a training target. And a constraint expressed
in the prompt is not a constraint: an instruction arm reading ``do not write in Chinese'' moved the
contamination rate by 0.0, while the same requirement imposed on the decoder moved it to zero.

The first SFT trained on 2{,}992 records and barely moved the model, because its targets were nearly
indistinguishable from the base model's own output. Rebuilding the corpus with the style axes enforced
at generation time and holding every hyper-parameter fixed, the second SFT trained on \emph{519}
records and produced the profile in Table~\ref{tab:style}.

All adapters are LoRA rank~8, $\alpha$=16, dropout~0, attached to every linear layer including the
\texttt{in\_proj} projections of the hybrid-attention blocks, and merged into the weights between
stages. \textbf{Style SFT} trains on the 519 style-rewritten records, lr $2\times10^{-5}$, one epoch, three
seeds; the run name seeds the data order, while LoRA $A$ initialisation is unseeded and therefore also varies between runs, so each seed is an independent draw of both but the initialisation is not reproducible from the seed alone. \textbf{Safety DPO} trains
2{,}757 preference pairs, $\beta$=0.15, effective batch 8, 345 steps; scenarios are seeded from KoSBi
categories mapped onto KoBBQ categories and written fresh by the teacher, which never sees a KoBBQ
item, and every training pair is checked against all 16{,}240 KoBBQ items by 12-character substring
match before training proceeds. \textbf{Compliance distillation} attaches a disclosure instruction to
the teacher, filters generated responses through a rule gate at a 0.478 pass rate leaving 1{,}205
records, holds out 129, and trains on the remaining 1{,}076 \emph{with the instruction removed}
(1{,}076 steps, final training loss 0.7278, held-out evaluation loss 0.5282, 26 minutes on one H200).
This is the FIN1 arm. \textbf{Both orderings} of the last two stages were trained, one run each: M4 is
style$\to$safety$\to$compliance, M5 is style$\to$compliance$\to$safety.

\textbf{Control arms.} The neutral arm replaces the style-rewritten target with the raw teacher response;
prompts verified identical, 519 of 519 matched, zero mismatches. The length-matched arm replaces it
with that same raw response truncated at a sentence boundary to the paired style target's character
length --- truncation only, never a rewrite, so every target is an exact contiguous prefix of its
neutral counterpart. Three seeds per control arm.

\subsection{What response-style alignment did to response style}

\begin{table}[t]\centering\small
\caption{Response-style profile over 637 held-out prompts, $T$=0.7, no system prompt, one serving stack.
Eleven deterministic features, each reported alone with its own paired bootstrap interval; we do not
combine them into a style score. Ten features are defined over all 637 paired responses; formal-register
share is reported conditionally on scoreable endings, with coverage shown separately. All reported
base--style contrasts exclude zero under their corresponding bootstrap estimand. Rates are percentages of responses; sentence, paragraph and length figures are
medians.}
\label{tab:style}
\begin{tabular}{lrrrr}\toprule
Feature & base & style & safety & compliance\\\midrule
Bullet rate & 97.3\% & 0.6\% & 0.0\% & 0.5\%\\
Heading rate & 85.1\% & 0.0\% & 0.0\% & 0.0\%\\
Length (characters) & $\ge$1{,}329$^{\dag}$ & 152 & 165 & 204\\
Length (tokens) & $\ge$700$^{\dag}$ & 77 & 85 & 99\\
Sentence count & 22 & 2 & 3 & 3\\
Paragraph count & 9 & 1 & 1 & 1\\
List items (mean) & 12.29 & 0.07 & 0.00 & 0.04\\
Closing-summary rate & 2.35\% & 0.47\% & 0.63\% & 0.94\%\\
Formal register share$^{\ddag}$ & 93.9\% & 50.9\% & 52.4\% & 89.5\%\\
Repeated-sentence rate & 1.26\% & 0.00\% & 0.00\% & 0.00\%\\
Non-Korean contamination & 9.11\% & 4.87\% & 20.72\% & 6.28\%\\\bottomrule
\end{tabular}
\\[3pt]{\footnotesize
$^{\dag}$ Right-censored: 74.6\% of base responses hit the 700-token generation cap, so base's length
entries are floors and every length gap against base is a lower bound. An uncensored re-run of the base
arm ($n$=150, cap 4096, 0 responses at the ceiling) gives a true median of 737 tokens and 1{,}478.5
characters.\\
$^{\ddag}$ Conditional on the response ending in a scoreable register marker; coverage differs by arm
(627, 448, 510 and 611 of 637 responses respectively), so this row compares like with like only within
each arm's scored subset, and the unscored fraction is itself part of the style difference.}
\end{table}

We operationalise response style through eleven deterministic output features rather than a learned
composite score. They were written for this task rather than derived from the rule engine
that produced the training targets, so the suite does not score the model on its own optimiser's metric; the
one deliberate reuse is the non-Korean-script regex, taken verbatim from the project's existing purity
check. The suite was required to discriminate a known-different pair before being used for comparison,
and it does: every feature separates base from the style arm with a paired bootstrap interval excluding
zero under its own estimand.

\subsection{Measurement}

KoBBQ all-test is stratified to at most 2{,}000 items per category, giving 16{,}240 KoBBQ items of which
8{,}139 are ambiguous and 8{,}101 disambiguated; answer options are rotated by a deterministic function
of the item identifier. Decoding is $T$=0, thinking off, \texttt{max\_tokens}=16.

The template identifier used as the clustering unit throughout is the item identifier's leading
\texttt{category-NNN} field, so that all lettered variants of a template form one cluster. One
provenance discrepancy is unresolved and we state it rather than smooth it. The published KoBBQ
release~\cite{jin2023kobbq} reports 268 templates over 76{,}048 samples; the copy of \texttt{all-test} we evaluate carries 269
template identifiers over 76{,}568 rows, of which 264 identifiers appear in the stratified ambiguous
subset. We have not identified which build ours came from. The discrepancy does not affect the
clustering: all 269 identifiers are well formed, per-category numbering is contiguous from 001 with no
gaps, the variant structure is regular --- 253 templates carry two lettered variants and 16, all in the
race, ethnicity and nationality category, carry four --- and no single template accounts for the 520
extra rows. The ambiguous evaluation subset covers 264 templates in 531 variants; the variant count is
not twice the template count because the per-category cap admits only part of some templates' variant
sets, which is why an odd number is possible at all. We treat the key as sound and the sample count as
not reconciled.

Disclosure is measured on 48 immutable prompts (16 solicitation, 16 product explanation, 8 plain
information, 8 unsuitable solicitation) at $T$=0.2, three runs per arm, and on a 192-prompt expansion
for the dose-response design of \S\ref{sec:dose}. The disclosure requirements are those of the Korean
Standard Investment Solicitation Rules~\cite{kofia2026sisr}, the self-regulatory model rule of the Korea
Financial Investment Association, provided for firms to adapt when establishing and operating their own
investment-solicitation rules; our rule wording is operationalised from its current revision. We
therefore treat the labels as benchmark disclosure requirements rather than as statutory obligations or
as estimates of legal compliance. Four of the five rules cite that current text directly. The fifth, that past returns do not
guarantee future returns, rests on an annex form for robo-advisory services together with a general
prohibition on misleading statements rather than on a positive disclosure duty in the general text; we
keep it in the instrument and record here that its basis is the narrower one.

Serving is vLLM 0.28.0, one GPU per arm --- B200 for the control wave of \S\ref{sec:rq1}, H200 for the
pipeline arms of \S\ref{sec:rq4}, with matching engine arguments: \texttt{max\_num\_seqs}=256, fp8 KV cache, prefix caching,
torch.compile enabled, speculative decoding with a DFlash drafter ($K$=7), \texttt{max\_model\_len}
32768. For the KoBBQ arms, serving arguments were read back from the live pod and diffed against
the reference arm --- the diff is empty apart from the two lines naming the model --- and before any
datum was recorded \texttt{/v1/models} was queried and the returned served-model name asserted against
the expected one. That assertion matters because a port-forward left pointing at another workstream's
endpoint returns HTTP 200 throughout and otherwise yields wrong-model data silently.

\textbf{We did not apply that assertion to the disclosure arms, and we correct the earlier version of
this paper, which claimed we had.} Those responses were generated by a separate harness that records
the model name from its own request configuration rather than reading it back from the endpoint, and
that harness stores no engine arguments. Local port-forwards were reused across arms: the same local
port carries one arm naming the base model and another naming the style checkpoint. Their run windows
are disjoint by nine minutes, which is what re-forwarding between arms looks like, but that is
circumstantial and the stored record cannot by itself establish which checkpoint served which arm. The
per-arm forwards no longer exist, so this cannot be verified after the fact. We report the disclosure
results as unverified on endpoint identity rather than restate a check we did not run.

All statistical tests are computed by one shared module, which decides from the stored item-identifier
sets rather than from the design as described which test is admissible: identical sets admit paired
tests, disjoint sets admit independent tests, and partially overlapping sets admit neither and fall
through to an overlap-aware bootstrap or a partition permutation. KoBBQ instantiates its items from a
smaller set of templates, so items sharing a template share their wording and are not independent: all
uncertainty estimates over KoBBQ items use the KoBBQ template as the resampling or clustering unit,
except where a statistic is explicitly described as an item-level diagnostic.

\section{RQ1: the untargeted behaviours move, and the target's form is why}\label{sec:rq1}

\subsection{Two arms are not enough}

That the style-aligned checkpoint differs from base on both untargeted axes is easy to measure and hard
to attribute. Any supervised fine-tune on this corpus might produce the same drift. So we built control
arms that hold the prompts, the recipe, the data volume, the holdout split and the serving
configuration fixed and change only the target text.

Together with the base model this is ten arms on one item set of 8{,}139 ambiguous items, with zero
unparseable completions and zero endpoint errors in every arm. All pairings are therefore classified as
paired with overlap fraction 1.000, so item-level McNemar diagnostics are computable throughout, and
because no arm drops an item all five exclusion conventions we considered give numerically identical
results.

\begin{table}[t]\centering\small
\caption{Ten arms on 8{,}139 ambiguous KoBBQ items and one serving configuration, read back from the
live pods. The base arm was re-evaluated on the arm serving configuration for this table; it parses
every item, so all ten arms score the same 8{,}139. $\Delta$ columns are against that matched base. The last two columns split $\Delta$ absolute stereotyped into the two terms of the exact
identity of \S\ref{sec:rq2}: the answer-propensity contribution $L_0\Delta A$ and the observed
composition residual $A_1\Delta L$. Every arm here scores all 8{,}139 items with zero exclusions, so
each contrast against base is on the full set; elsewhere a contrast is computed on the item set the two
arms share, so $n$ differs by pair. Median length is over the 637 held-out prompts of
Table~\ref{tab:style} and is right-censored on every row except the three style seeds, where under
1.3\% of responses reach the generation cap against 55--75\% elsewhere.}
\label{tab:tenarm}
\begin{tabular}{lrrrrrr}\toprule
& Answer & $\Delta$ answer & $\Delta$ abs. & Answer-propensity & Observed comp. & Median\\
Arm & rate & rate & stereo. & contribution & residual & len.\\\midrule
base            & 18.59\% & ---      & ---      & ---      & ---      & 1{,}329\\\midrule
style s1        & 19.93\% & $+1.339$ & $+1.241$ & $+1.154$ & $+0.087$ & 152\\
style s2        & 19.04\% & $+0.455$ & $+0.565$ & $+0.392$ & $+0.173$ & 74\\
style s3        & 19.24\% & $+0.651$ & $+0.663$ & $+0.561$ & $+0.102$ & 118\\\midrule
length-matched s1 & 18.01\% & $-0.577$ & $-0.430$ & $-0.498$ & $+0.068$ & 1{,}305\\
length-matched s2 & 18.58\% & $-0.012$ & $+0.061$ & $-0.011$ & $+0.072$ & 1{,}300\\
length-matched s3 & 17.72\% & $-0.872$ & $-0.749$ & $-0.752$ & $+0.002$ & 1{,}314\\\midrule
neutral s1      & 17.20\% & $-1.388$ & $-1.167$ & $-1.197$ & $+0.029$ & 1{,}292\\
neutral s2      & 16.99\% & $-1.597$ & $-1.302$ & $-1.377$ & $+0.074$ & 1{,}298\\
neutral s3      & 16.99\% & $-1.597$ & $-1.278$ & $-1.377$ & $+0.099$ & 1{,}293\\\bottomrule
\end{tabular}
\end{table}

\subsection{The sign reverses}

Against the matched base the style arms raise the answer rate by $+1.339$, $+0.455$ and $+0.651\pp$ and
the neutral arms lower it by $1.388$, $1.597$ and $1.597\pp$. The two observed seed ranges do not overlap ---
the style minimum ($+0.455$) exceeds the neutral maximum ($-1.388$) --- and the mean gap of $2.343\pp$
is $6.9\times$ the pooled within-arm standard deviation of $0.339\pp$ and larger than either arm's own
seed range (0.885 and $0.209\pp$). At the item level, McNemar tests give the same direction in all of
these comparisons --- $|z|$ from 2.52 to 9.64 against the matched base and 9.76 to 14.63 across the nine
style-versus-neutral pairings --- but they treat the items as independent and ignore KoBBQ's template
dependence, so we report them as checkpoint-conditional diagnostics and take the template bootstrap
below as the inference. The mean gap is a difference between two arms and does not depend on which base they are
measured against; the individual $\Delta$ values do, which is why the base was re-evaluated on the arm
serving configuration before this section was written.

Those item-axis tests treat the 8{,}139 items as independent, and they are not: KoBBQ instantiates many
items from a smaller set of templates, and the items sharing a template share their wording. We
therefore recomputed the contrasts with a cluster bootstrap that resamples \emph{templates} --- the
identifier prefix, so that all lettered variants of one template form one cluster --- over the 264
templates the ambiguous split covers, 20{,}000 draws. The between-arm claim survives intact: all nine
style-versus-neutral cross-pairings exclude zero on both answer rate and absolute stereotyped exposure,
and the mean gap is $+2.343\pp$, CI $[+1.177, +3.600]$, on answer rate and $+2.072\pp$, CI $[+1.020,
+3.216]$, on absolute exposure. The per-arm contrasts against base are weaker than the item-axis tests
made them look. Style seed 1 excludes zero ($+1.339\pp$, CI $[+0.269, +2.577]$), but \textbf{style seeds
2 and 3 cover zero} --- $+0.455\pp$, CI $[-0.513, +1.437]$ and $+0.651\pp$, CI $[-0.424, +1.858]$ --- as
does the length-matched second seed, while all three neutral arms exclude zero. Resampling the 531
lettered variants instead narrows the intervals without changing any of these verdicts. So a single
style seed's shift against base is not established on the item axis at this clustering, and we do not
claim it is. What is established is the between-arm separation, and the generalisation this section
rests on is across the three training runs per arm rather than across items. We report the item-level
McNemar statistics as diagnostics conditional on each checkpoint.

The three-against-three exact permutation test returns $p$=0.100, which needs reading rather than
dismissing. With $\binom{6}{3}=20$ assignments the two-sided floor \emph{is} 0.100, so a $p$ at the
floor means the groups separated completely and a $p$ above it means they did not. The strength here
comes from three things a $p$-value cannot carry: the sign reverses, the observed seed ranges do not
overlap, and the template-clustered intervals exclude zero in all nine style-versus-neutral
cross-pairings. With $n$=3 per arm we report the observed seed range for every arm and a standard
deviation only where it is used in a comparison, and lean on neither as an estimate of training-run
variability.

So the direction is not a property of fine-tuning on this corpus. It is a property of the form of the
training target: the neutral, length-matched and style arms are ordered monotonically by their mean
effects.

\subsection{Which property of the form, we do not know}

The style and neutral arms differ on two axes at once: the style targets are short (median 143
characters) and the raw ones are long (median 1{,}076). The length-matched arm was built to separate
them under a rule pre-specified before any evaluation number existed --- inside the style range means
length is the lever, inside the neutral range means the style rewrite is, strictly between means the
design does not resolve it at $n$=3.

It landed strictly between: $-0.577$, $-0.012$ and $-0.872\pp$, mean $-0.487\pp$. We honour the rule as
written. Neither ``style alignment causes the shift'' nor ``target length causes it'' survives as a
clean single-lever claim. What the arm does establish is that it separates from both --- $1.040\pp$
from neutral and $1.302\pp$ from style, non-overlapping in each direction --- so it rules out either
original target form as a complete explanation. It does not go further: because truncating to a length
target also strips clarifying questions, hedges and enumerations, as the next paragraph measures, the
arm does not identify separable contributions of target length and the broader style rewrite.

Two secondary checks pre-specified in advance both fired, in opposite directions. The first does not
support a simple transfer-through-generation-length mechanism: if short training targets taught the model to write short answers, and short answers left
less room to hedge, this arm's generations should be short. They are not --- median 1{,}305, 1{,}300 and
1{,}314 characters, within 12 characters of the neutral arm --- so training-target length did not
transfer to generation length at all. The second undermines the arm's own label: truncating a raw
response at a sentence boundary produces the response's \emph{opening}, which is the part that commits,
and deletes the part that asks and qualifies. Over the 519 targets, clarifying questions fall from
33.7\% to 8.7\%, requests for more information from 25.4\% to 7.1\%, hedges from 51.5\% to 23.5\% and
markdown enumerations from 58.0\% to 15.2\%, most landing near the style arm's own rates. The arm is
length-matched to the style target (per-record Pearson $r$=0.9964, median absolute residual 16 characters) and not
discourse-matched. That the manipulation reached discourse-level features is also why we call the
intervention response-style rather than surface-style: it is not a surface-only change.

\subsubsection*{The fourth cell}

That leaves the cell the three arms do not cover: a target that is short \emph{and} preserves hedges and
clarifying questions. We built it, gated it, and trained it at three seeds on the same recipe and the
same 8{,}139 items. The construction gate passed on all six checks against thresholds committed before
the data existed; a first attempt failed one of them and we changed the construction rather than the
threshold. The built targets carry hedges in 59.9\% of records, a question in 32.2\% and a request for
more information in 47.8\%, at a median of 153 characters. Unlike the length-matched arm, the short
target transferred: median generated response 195 characters, against 115 for style, 1{,}294 for neutral
and 1{,}306 for length-matched. The cell is genuinely short-and-hedging.

\begin{table}[t]\centering\small
\caption{All four target-form arms, three training seeds each, change in ambiguous answer rate against
the matched base on the same 8{,}139 items. The seed range is what separates the fourth arm from the
other three; Appendix~\ref{app:fourth} gives the pre-specified decision rule and why it cannot
discriminate at this dispersion. That rule's landmarks were fixed before the base was re-evaluated, so
the classification is reported as specified while the deltas here are on the matched base; the shift is
a constant and does not change which landmark each seed falls nearest.}
\label{tab:fourarm}
\begin{tabular}{lrrrrr}\toprule
Arm (target form) & seed 1 & seed 2 & seed 3 & mean & range\\\midrule
neutral (long, hedging)              & $-1.388$ & $-1.597$ & $-1.597$ & $-1.528$ & 0.21\\
length-matched (short, low hedging)  & $-0.577$ & $-0.012$ & $-0.872$ & $-0.487$ & 0.86\\
style (short, low hedging, rewritten)& $+1.339$ & $+0.455$ & $+0.651$ & $+0.815$ & 0.88\\
hedge-preserving (short, hedging)    & $-0.639$ & $+1.708$ & $-2.519$ & $-0.483$ & \textbf{4.23}\\\bottomrule
\end{tabular}
\end{table}

It does not identify the responsible feature. The three seeds are $-0.639$, $+1.708$ and $-2.519\pp$
(Table~\ref{tab:fourarm}), a range of $4.23\pp$ against $0.21$--$0.88$ for the other three arms. That
dispersion is too large for any location statistic computed on it to discriminate the hypotheses: the
three seeds fall nearest three different landmarks, and every pairwise contrast involving the arm
overlaps. Appendix~\ref{app:fourth} gives the pre-specified rule, why it was mis-specified for an arm
this dispersed, and the accompanying detail.

The control also moved more than the feature it was built to isolate. A 150-character answer that keeps
the teacher's clarifying question usually has to end on it, and this arm ends on a question in 15.2\% of
targets against 3.1\% for style and 4.6\% for neutral, so hedging and turn-final structure vary together
exactly as truncation varied length and discourse form together. Separately, the generator for this arm
saw a system prompt carrying up to three sentences of raw teacher text and the style arm's did not, which
is a different conditioning distribution rather than a different instruction. Both differences are
uncontrolled here.

We therefore do not write that response-style alignment in general raises answer propensity. The
deployed checkpoint is the one trained on short style-rewritten targets, and for pipeline accounting the
attribution to that stage is correct; the generalisation is not yet available.

One thing this design cannot do is measure length as a dose. Across the six style and neutral
checkpoints the change in median generation length and the change in answer rate correlate at
$r=-0.965$, but 1{,}140 of the $x$-axis's 1{,}224-character span (93.1\%) is a single gap between the
two clusters, so that coefficient is the between-arm mean difference rewritten, not a dose-response.
Within the style arm the sign is the opposite of the prediction. Assigning the dose is what
\S\ref{sec:dose} does on the other axis.

\subsection{The second axis moved too}\label{sec:rq1-compliance}

On the four arms that share one serving stack, over 456 (response, required-rule) pairs per arm, the
strict detector marks disclosure in 27.6\% of pairs for base, 6.1\% for the style-only arm, 94.1\% for
the prompted arm and 94.5\% for the compliance-distilled arm; a second, independently specified looser
detector marks 36.4\%, 14.0\%, 98.5\% and 95.2\%. Both instruments register a large drop at the style
stage and a large recovery at the distillation stage. These are detector-identified disclosure rates,
not estimates of regulatory compliance: they measure how often the required language is emitted in a
form the instrument recognises, and \S\ref{sec:noninvariance} shows the instruments do not recognise it
equally across arms.

\section{RQ2: absolute exposure changes are dominated by answer propensity}\label{sec:rq2}

\subsection{The decomposition is an identity; the allocation is the result}

Write $A$ for the probability that the model answers an ambiguous item, $L$ for the probability that an
answer is the stereotyped option, and $S = A \cdot L$ for absolute stereotyped exposure. Then for any
two arms, exactly and without approximation,

\[
\Delta S \;=\; L_0\,\Delta A \;+\; A_1\,\Delta L .
\]

This is algebra. Nothing about a model can make it fail, and the last two columns of
Table~\ref{tab:tenarm} are its two terms rather than a prediction and an error --- we verified that the
observed-composition column equals $A_1\Delta L$ to within floating-point tolerance on
every arm. The smallness of that column is therefore not an empirical law fitted to within a band, and
we do not report it as one.

The empirical content is where the movement went, and we state it in percentage points before any
ratio. Restricting at the outset to the seventeen contrasts produced by training --- the twelve control
checkpoints of Tables~\ref{tab:tenarm} and~\ref{tab:fourarm} and all five pipeline stages of
Table~\ref{tab:order}, sixteen distinct checkpoints in all, since the style checkpoint appears in two
tables as its two runs --- the answer-propensity term $L_0\Delta A$ ranges from $-13.42$ to
$+1.47\pp$, while the composition term $A_1\Delta L$ stays between $+0.00$ and $+0.45\pp$ --- every one
of the seventeen is positive and under half a point. The three seeds of the unstable fourth arm are
inside that pattern rather than an exception to it: their answer-propensity terms are $-0.55$, $+1.47$
and $-2.17\pp$, spanning more than any other control arm and supplying the upper end of the span above,
while their composition terms are $+0.15$, $+0.09$ and $+0.27\pp$. What that arm destabilises is the
magnitude and sign of the answer-propensity shift, not the observed allocation between the two terms. Adding the two precision arms below, which were
measured for another purpose and are a generalisation check rather than an experiment we ran for this
question, extends the propensity span to $+2.88\pp$; their composition terms are $-0.03$ and $+0.06\pp$,
the same order of magnitude, and the NVFP4 arm's is the one negative value we observe.
The largest composition terms are $+0.445\pp$ for the safety stage and $+0.445\pp$ for M4 --- both
stages that raise abstention sharply --- against answer-propensity terms of $-9.56$ and $-8.85\pp$ on
the same contrasts. The two remaining pipeline stages sit inside the same pattern: FIN1 has a
composition term of $+0.220\pp$ against a propensity term of $-5.53\pp$, and M5 $+0.123$ against
$-13.42\pp$. Nothing forced that: a treatment could have moved $L$ enough to reverse the sign of
$\Delta S$ relative to $\Delta A$, or to double it, and in this pipeline none did.

As a secondary summary, the share of $\Delta S$ carried by answer propensity runs from 93 to 116
percent. That ratio is a quantity only where $\Delta S$ is itself resolvable, and under the template
clustering of \S\ref{sec:setup} that is thirteen of the seventeen training contrasts: the two smaller
style seeds, the second length-matched seed and the first seed of the fourth arm all have
template-clustered intervals covering zero, and on those the denominator has collapsed rather than the
composition term having grown. The W4A16 arm is in the same position. We quote the ratio only where
$\Delta S$ resolves; the percentage-point statement above does not depend on that restriction, because
a term is a term whether or not the difference it decomposes is distinguishable from zero.

Contrasts are computed on the item set the two arms share, pairwise, rather than on an intersection
across every arm in a table --- otherwise a two-arm comparison would depend on which other arms happen
to be present. The five pipeline contrasts quoted here are on $n$=8{,}139 (base to style), 7{,}991
(base to safety), 8{,}139 (base to FIN1), 8{,}122 (base to M4) and 8{,}139 (base to M5); the matched base
drops no item, so only the safety and M4 arms narrow their pairings.

\textbf{Because the conditional term is evaluated on treatment-dependent answered subsets, we do not
interpret the residual as evidence that latent stereotype preference is invariant.} $\Delta L$ compares
the stereotyped share of two different sets of answered items, and which items those are is itself an
outcome of the treatment. \S\ref{sec:selection} makes that argument properly. The claim here is
narrower and is about where a deployment-visible quantity comes from: what a user encounters is $S$,
and in this pipeline $S$ moved because $A$ moved.

Two further observations bear on how far that travels.

\textbf{It runs in both directions.} The neutral arms answer \emph{less}, and their absolute stereotyped
exposure falls, again dominated by the answer-propensity term. An allocation that held only for
increases would be a weaker observation than one that also covers decreases produced by a different
training target.

\textbf{It is not confined to training interventions.} Serving one checkpoint at three weight precisions
with matching values for the four serving knobs the ledger records --- KV-cache dtype, maximum sequence
count, speculative decoding and memory utilisation, with no argument readback stored for these arms and
different kernel paths for the two quantizers --- on the 7{,}867 ambiguous items all three arms parsed so that
every contrast is fully paired, the NVFP4 arm raises the answer rate by $3.11\pp$ against bf16
(template-clustered CI $[+1.70, +4.83]$) and absolute stereotyped exposure by $2.86\pp$ (CI $[+1.50,
+4.50]$), with an answer-propensity term of $+2.88\pp$ and a composition term
of $-0.03\pp$. A W4A16 arm moves neither quantity ($-0.29\pp$, CI $[-1.03, +0.46]$; $-0.23\pp$, CI
$[-0.97, +0.52]$), with a composition term of $+0.06\pp$. That compression moves bias measurements
is itself reported: \citet{rath2026quantization} find 3-bit quantization drives 6--21\% of previously
unbiased BBQ items into stereotypical behaviour while perplexity moves under 0.5\% at 8-bit, and
\citet{hua2026quantized} find quantization flips bias on up to 21\% of items with aggregate bias scores
unchanged, correlated with model uncertainty. Both report item-level bias changes under quantization
associated with uncertainty or with changes in unknown-option selection, which is prior evidence of a
similar measurement channel; we have not shown that the mechanism is the same one. These arms are a
single run each with unmeasured requantization build spread, so we use them only as a generalisation
check and never as an effect size; they are also why a
contrast crossing weight precision cannot be read as an alignment contrast.

\subsection{The magnitude is a range}

Three style seeds give absolute stereotyped increases of $+1.241$, $+0.565$ and $+0.663\pp$ against the
matched base (Table~\ref{tab:tenarm}); only the first has a template-clustered interval excluding zero,
so these are point estimates and the two smaller ones are not individually established. The spread of $0.676\pp$ between seeds is larger than the smallest seed's whole effect. For scale,
re-evaluating one checkpoint in the same session on the same node and serving spec moved its absolute
stereotyped exposure by $0.013\pp$, template-clustered CI $[-0.151, +0.136]$, not distinguished; the
same checkpoint measured once on an H200 and once on a B200 differed by $0.098\pp$, likewise not
distinguished. The first is
a within-session reproducibility statement, the second a cross-hardware observation, and neither is an
estimate of variance across training runs.

So a single training run's value cannot be written as the effect. The defensible statement is that
across the three style seeds the point estimates for absolute stereotyped exposure increase by
$0.57$--$1.24\pp$ relative to the matched base, with the point-estimate direction the same at every seed
and the magnitude varying $2.2\times$. Three points have a standard deviation; what they do not have is
a stable estimate of variability across training runs, and it is that estimate the range is standing in
for. The alignment step itself is
resolvable: base to style is $+1.143\pp$ on an identical item set of 8{,}139, template-clustered CI
$[+0.16, +2.27]$, with an answer-propensity term of $+1.133\pp$ and a composition term of $+0.010\pp$.

\subsection{On the compliance axis, the emission variable is length}\label{sec:dose}

The style-aligned checkpoint's free-running disclosure gap against base is $-22.25\pp$ under the loose
detector (CI $[-26.56, -17.94]$) and $-21.83\pp$ under the strict gate (CI $[-26.44, -17.25]$), pooled
over 456 rule-pairs per arm with prompt-level clustering and overwhelmingly significant under bootstrap,
GEE and GLMM. The same stage shortens responses by an order of magnitude (Table~\ref{tab:style}). The
obvious hypothesis is that the second explains the first.

An observational version of that test --- regressing disclosure on arm with $\log$ length as a
covariate --- appeared to support it and did so too strongly, flipping the arm coefficient's sign. We do
not use that result. Length is a post-treatment variable and conditioning on it over-adjusted. The
correct design assigns the dose.

We did that with a neutral length instruction in the system prompt at four targets (100, 200, 400 and
800 characters) across three arms, over an expanded 192-prompt set giving 608 rule-pairs per cell and
7{,}296 record pairs in total, at $T$=0.2 with a generous token budget --- \texttt{max\_tokens} was
never the length lever, because truncation would amputate a sentence-final disclosure and manufacture
the effect under test, and the \texttt{finish\_reason=="length"} count is zero in every cell. Arms ran
sequentially on one B200 with byte-identical serving arguments apart from the model name. We report
intention-to-treat by assigned target rather than by realised length, for the same reason we abandoned
the covariate adjustment.

\begin{table}[t]\centering\small
\caption{Detector-identified disclosure by assigned length target, intention-to-treat, in percentage
points against base, with 95\% bootstrap intervals resampled over the 160 prompts that contribute at
least one applicable disclosure rule pair --- the full set is 192, of which the 32 plain-information
prompts carry no required rule --- with 20{,}000 draws clustered on prompt rather than on rule-pair. Each cell is 608 rule-pairs. Bold marks intervals
excluding zero.}
\label{tab:dose}
\begin{tabular}{lrrrr}\toprule
& \multicolumn{2}{c}{Style-only arm vs base} & \multicolumn{2}{c}{Distilled arm vs base}\\
\cmidrule(lr){2-3}\cmidrule(lr){4-5}
Assigned target & Loose & Strict & Loose & Strict\\\midrule
100 characters & $-1.97$ & $-0.99$ & $\mathbf{+59.7}$ & $\mathbf{+61.5}$\\
               & \footnotesize$[-4.12,+0.16]$ & \footnotesize$[-2.66,+0.66]$ & \footnotesize$[+54.7,+64.5]$ & \footnotesize$[+55.9,+66.9]$\\
200 characters & $+0.66$ & $-0.33$ & $\mathbf{+70.7}$ & $\mathbf{+76.5}$\\
               & \footnotesize$[-1.62,+2.88]$ & \footnotesize$[-2.30,+1.62]$ & \footnotesize$[+66.2,+75.0]$ & \footnotesize$[+72.1,+80.9]$\\
400 characters & $\mathbf{-5.10}$ & $\mathbf{-3.29}$ & $\mathbf{+71.2}$ & $\mathbf{+82.1}$\\
               & \footnotesize$[-7.80,-2.45]$ & \footnotesize$[-5.28,-1.31]$ & \footnotesize$[+67.5,+74.9]$ & \footnotesize$[+78.5,+85.5]$\\
800 characters & $\mathbf{-5.26}$ & $\mathbf{-4.44}$ & $\mathbf{+71.7}$ & $\mathbf{+81.9}$\\
               & \footnotesize$[-8.37,-2.20]$ & \footnotesize$[-7.00,-1.97]$ & \footnotesize$[+68.4,+75.0]$ & \footnotesize$[+78.8,+84.9]$\\\bottomrule
\end{tabular}
\end{table}

Most of the style stage's detector-measured disclosure gap is reproduced by assigned length. Free-running
it is about 22 points; with length assigned it is 3 to 5 points at the two longer targets, where the
prompt-clustered intervals exclude zero under both detectors, and indistinguishable from zero at the two
shorter ones, where every interval covers zero. But the curves do not coincide: 3 to 5 points survive at
400 characters and above under both detectors. The correct statement is that most of the
detector-measured disclosure gap is reproduced by assigned verbosity, which establishes that response
length is \emph{sufficient} to reproduce most of the gap and does not identify it as the causal mediator
through which style post-training produced the original change. That is why the table is
indexed by assigned target and not by realised length.

The compliance-distillation stage does not work by lengthening the response. Its advantage over base is $+59.7$ to
$+82.1\pp$ at every assigned target under both detectors, and at the 800-character target its median
response is 620 characters against base's 1{,}062 --- it clears the bar by a wide margin while writing
barely more than half as much. The observational analysis agrees on this arm, where the length-adjusted
coefficient rises rather than falls; it is a log-odds coefficient and is not on the same scale as the
assigned-dose percentages, so the two results support the same direction independently rather than
calibrating each other.

\subsection{Disambiguated-item accuracy as a task-local control}

An emission account would read differently if the same task's evidence-given condition had collapsed
alongside it. It did not. On KoBBQ's disambiguated items --- where the context does support an answer ---
accuracy ranges from 86.24\% to 89.82\% across the evaluated checkpoints against 88.20\% for base
(Table~\ref{tab:order}). We report this descriptively: some stages sit above base and others below. It
is a task-local control on the same benchmark and speaks to nothing outside it.

\section{RQ3: evaluation changes with the output distribution}\label{sec:rq3}

Alignment work of this kind is measured with instruments built before the treatment existed. When the
treatment's content is a change in the surface distribution of outputs, whether those instruments
survive it is an empirical question --- and one already answered in part for bias benchmarks, where
\citet{merzlyakova2026format} show that varying the answer format across closed, Likert and open-ended
conditions moves measured gender bias enough to reverse rankings, with refusal as the channel. Two of
ours did not survive, and both failures are results rather than housekeeping. Defects that changed a number but not a conclusion are in Appendix~\ref{app:instruments}.

\subsection{Conditional-share contrasts are selection-confounded}\label{sec:selection}

The most widely used quantity on a paired bias benchmark is the share of \emph{answered} items that are
stereotyped, and the conditional bias score is a monotone transform of it. That share is perfectly
measurable, and we measure it. What it does not do is identify what it is usually read as identifying:
\textbf{a between-arm contrast in conditional stereotyped share does not identify a change in
conditional content preference when answer status is treatment-dependent.}

The argument is short. The share is defined over each arm's answered subset. When the treatment changes
which items get answered, the answered subset is itself an outcome of the treatment, so comparing shares
across arms conditions on a post-treatment variable. It is a selection effect, not a parsing problem,
and careful reporting does not repair it.

It is measurable how large the selection is. Between base and the style arm --- two arms with
\emph{zero} unparseable completions on either side --- the answered sets overlap at a Jaccard index of
0.860. Between the safety arm and the style arm it is 0.346, the value Table~\ref{tab:transition}
implies; between the style arm and the compliance arm, 0.603. The two arms in each pair are not scoring the same
questions, and the difference between the question sets \emph{is} part of the treatment effect.

The consequence is that the null distribution is not centred at zero. In this regime, permuting arm
labels alone produces a non-zero share gap purely from composition. When we tested our own earlier
claim that the safety stage raises the conditional share, an item-level partition permutation put the
null between $+4.73$ and $+7.04\pp$ and contained the observed value at $p$=0.87. That permutation
treats the items as independent, so like the McNemar statistics it is a diagnostic and not one of this
paper's inferential results; we report it as evidence that arm-label exchangeability alone already
produces a gap of the observed size, and the claim of this section does not rest on it. The claim is
structural, and holds whatever a resampling test returns: when answer status is treatment-dependent,
the contrast does not identify what it is read as identifying. Of the six conventions we considered for the one arm that drops
items, the two defensible ones --- dropping the unparsable items, and recovering their raw completions
and scoring what they actually said --- both fail to detect the effect. A third convention, scoring
every unparsable item as counter-stereotyped, is superseded: recovering the raw text showed that of the
42 of 148 that had in fact chosen an option, 34 were stereotyped and 8 counter-stereotyped, so the
assumption behind it is the reverse of what the outputs show. We drop it from the live comparison.

\begin{table}[t]\centering\small
\caption{Item-level transitions from the style checkpoint to the safety checkpoint, over the 8{,}139
ambiguous items both score. Rows are the style arm's outcome, columns the safety arm's. The safety
stage removes 827 of the style arm's 1{,}397 stereotyped answers and adds 11; the 517 that survive are
96.6\% of the safety arm's 535 stereotyped answers. Counts here are over all 8{,}139 items, including
the unparseable column; rates elsewhere in the paper use the scored denominator, so 535 appears as
6.70\% in Table~\ref{tab:order} and as 6.57\% of 8{,}139 here. The marginals of this matrix are the
style arm of Table~\ref{tab:tenarm} (6{,}519 / 1{,}397 / 223) and the \textsc{safety} checkpoint of
Table~\ref{tab:order} (535 stereotyped).}
\label{tab:transition}
\begin{tabular}{lrrrrr}\toprule
style $\downarrow$ / safety $\rightarrow$ & abstain & stereotyped & counter & unparsable & total\\\midrule
abstain      & 6{,}421 & 11  & 0  & 87 & 6{,}519\\
stereotyped  &    827  & 517 & 10 & 43 & 1{,}397\\
counter      &    168  &   7 & 30 & 18 &    223\\\midrule
total        & 7{,}416 & 535 & 40 & 148 & 8{,}139\\\bottomrule
\end{tabular}
\end{table}

Table~\ref{tab:transition} shows the selection happening item by item, and it is the strongest support
we have for this section's claim. Going from the style checkpoint to the safety checkpoint, the
conditional stereotyped share rises. Almost none of that rise is new stereotyped answering: the safety
arm adds 11 stereotyped answers to items the style arm abstained on, and removes 827 of the 1{,}397 the
style arm had. What is left is a residue --- 517 items, 96.6\% of the safety arm's 535 stereotyped
answers, every one of them already stereotyped before the safety stage. The share rises because the
intervention disproportionately removes items that had contributed to the denominator, leaving a
concentrated subset of previously stereotyped answers. That is a fact about which items remain in the
denominator, not about what the model prefers on a fixed set of items.

We report conditional shares in Table~\ref{tab:order} for completeness and use none of them as evidence
about content preference. This is also why \S\ref{sec:rq2} stops at an allocation statement about $S$: a
small composition term is what a treatment that changes answer propensity produces mechanically, and it
cannot be turned into a claim about what the model latently prefers.

\subsection{Disclosure-detector agreement is not invariant across treatments}\label{sec:noninvariance}

We have two detectors for required disclosure. The strict one implements the five rules of
\S\ref{sec:setup} as regular expressions over the wording of the Standard Investment Solicitation
Rules~\cite{kofia2026sisr}, with the basis qualification recorded there; the
loose one uses broader per-rule patterns, and the two were specified independently. Both are
reproducible. Neither is a measurement of regulatory compliance, and this section does not need it to
be: the result here is about the \emph{relationship between the two instruments}, which is observable
without reference labels of any kind.

\begin{table}[t]\centering\small
\caption{Agreement between the two disclosure detectors, by arm, over the 456 (response,
required-rule) pairs each arm contributes on the shared serving stack. Counts only; no reference
labels enter this table. Conditional agreement is $P(\text{strict}=1 \mid \text{loose}=1)$, i.e.\
``both'' divided by ``loose $+$''.}
\label{tab:detectors}
\begin{tabular}{lrrrrrrr}\toprule
Arm & Strict $+$ & Loose $+$ & Both & Strict only & Loose only & Neither & Cond.\ agreement\\\midrule
base            & 126 & 166 & 112 & 14 & 54 & 276 & 0.675\\
style-only      &  28 &  64 &  28 &  0 & 36 & 392 & 0.438\\
prompted        & 429 & 449 & 427 &  2 & 22 &   5 & 0.951\\
distilled       & 431 & 434 & 428 &  3 &  6 &  19 & 0.986\\\bottomrule
\end{tabular}
\end{table}

Table~\ref{tab:detectors} is the result. Conditional agreement between the two instruments --- the
fraction of loose-positive rule-pairs the strict gate also marks positive --- runs 0.675 on base, 0.438
on the style-only arm, 0.951 on the prompted arm and 0.986 on the distilled arm, a spread of 54.9
points. Which instrument you use therefore matters far more on some checkpoints than on others, and
that is a property of the treatments, not of a labelling protocol. On an earlier, independent set of
rollouts the same quantity spans 27.9 points, so this is not an artefact of the newer run.

The disagreement is directional rather than random, and the counts show where it lives. On base, 13 of
the 14 pairs the strict gate catches and the loose one misses are the investor-responsibility rule,
where the model writes the canonical clause in a word order the loose pattern cannot match. On the
style-only arm the strict-only cell is exactly zero and the loose-only cell is 36, so on that arm the
looser instrument is doing all of the additional marking. Median response length differs about
thirteenfold across these arms, so each detector's behaviour is entangled with the arm's verbosity ---
the same channel \S\ref{sec:dose} manipulates.

We had hypothesised that verbosity was therefore the \emph{mechanism} of the non-invariance. Assigning
length within a single arm does not support it: with the same prompt clustering \S\ref{sec:dose} uses,
conditional agreement is flat in length for base ($+0.06$, CI $[-0.19, +0.31]$) and for the style arm
($-0.05$, CI $[-0.36, +0.25]$), rising only for the distilled arm ($+0.79$, CI $[+0.25, +1.34]$). Pooling arms reverses the sign, which makes any cross-arm
trend a mixing artefact. The emission result and this one are parallel; neither produces the other.

The 18-row fixture set on which the strict gate was adopted is a unit test rather than a validation: it
holds sentences the author wrote to make the patterns fire and the same sentences with the phrasing
deleted. It says nothing about how either instrument behaves on real output, which is why the claim in
this section is built from the joint distribution of the two detectors and not from either one's
accuracy.

\subsection{The pre-specified abstention verdict is format-sensitive}\label{sec:rotation}

That option order moves multiple-choice results is established.
\citet{pezeshkpour2024order} report performance gaps of approximately 13\% to 85\% under option reordering and
attribute them to positional bias arising where the model is uncertain between its top choices, and
\citet{zheng2024selectors} identify a selection bias toward particular option IDs, trace it to token
bias, and propose a label-free inference-time correction. We replicate that instability and extend it
in one direction: to a \emph{pre-specified pass/fail verdict} on a \emph{social-bias} benchmark, where
the unit that moves is the choice between answering and abstaining rather than the choice between two
substantive options. We measure it on the shipped safety checkpoint, holding weights, decoding and the
item set fixed and varying only where the answer options sit.

The primary design is a fully paired cyclic panel: every item in the panel is scored under all three
cyclic rotations, so option position and item difficulty are separated by construction. The panel is a
separate draw of 250 items per category from the test set, restricted to ambiguous items
\emph{afterwards} rather than drawn from them, under sampling seed 20260909; because every category in
this build is exactly half ambiguous, that leaves 1{,}506 items at roughly 125 per category (observed
112 to 139), and the analysis keeps the 1{,}402 whose complete cycle came back scored (4{,}206
responses). Dropping an item
whose cycle is incomplete was fixed in the scorer before any arm was run and is a requirement of the
design rather than a filter applied to results: a partial cycle no longer visits every slot equally and
would manufacture the very slot effect being measured. The 104 dropped items are counted and
reported. Two effects appear. The chosen
slot is not uniform --- 31.8 / 34.1 / 34.2\% against 33.3 expected, with template-clustered intervals
excluding uniform on all three slots. One scorer defect bears on this figure and was found after the
first version of this paper: the letter parser accepted adjacent-letter and punctuation-only completions
(``AB'', ``(AB)'', ``.'') as the first slot instead of excluding them. It is fixed in the released code.
Its incidence cannot be recovered from the stored outputs, which retain raw completions only for excluded
items; every such completion would have been counted toward the first slot, which is the under-chosen
slot in every arm, so the correction can only deepen the deficit reported here, and under rotation any
such item is a uniform draw over the three options rather than a directional error. And abstention itself depends on where the ``unknown'' option
sits: 92.58\% when it is in the first slot against 95.36 and 95.22\% in the other two. Because the
complete cycle gives every item one response per slot, that comparison is paired within item, and the
template-clustered bootstrap on the paired differences is what carries it: $+2.78\pp$, CI $[+0.87,
+5.15]$, for the second slot against the first and $+2.64\pp$, CI $[+0.56, +5.30]$, for the third. A
random-intercept logistic fit on the same data, grouped by template rather than by item because items
sharing a template are correlated, puts the slot dummies at $z$=3.85 and 3.64 in the same direction ---
but the penalised quasi-likelihood fitter does not report convergence for either the item or the
template specification, so we report those coefficients as descriptive of the fitted model rather than
as calibrated tests, and rest the conclusion on the bootstrap. In aggregate a single
fixed rotation misstates position-neutral abstention by only $0.21\pp$ and the conditional share by
$-1.94\pp$ --- but at the item level, \textbf{7.8\% of items flip between abstaining and answering on
option position alone}.

A secondary check varies the per-item rotation through ten independent salted hashes over the full
8{,}139-item set. Abstention comes out at 92.81\% on average with a standard deviation of $0.17\pp$,
spanning 92.60 to 93.16\%. Our pre-specified criterion was $\ge$93\%. The conditional stereotyped share
is directionally stable across the same ten (93.72--95.77\%, every one above the value we report), so
what moves here is the threshold verdict and not the direction.

The finding is not that option position matters --- that is
\citet{pezeshkpour2024order} and \citet{zheng2024selectors} --- nor that the 93\% line is the wrong
line. It is that this known instability is large enough near a decision boundary to move a
pre-specified verdict on a social-bias benchmark: \textbf{the same checkpoint fails under eight of ten
defensible rotations and passes under two.} We report the criterion as format-sensitive rather than as
met. The cyclic panel is also what a
rotation-robustness claim has to be built on: the same items under all rotations, with no seed involved.

\section{RQ4: composition and order}\label{sec:rq4}

Both orderings of the last two stages were trained, one run each. This section is descriptive. Order
effects in sequential post-training are well
covered~\cite{ung2024chained,bhandari2026beyond,cha2025alignment,chen2024order,sweeney2026liebracket,fernando2024forgetting,liu2026representation},
we make no novelty claim, and with $n$=1 per ordering we cannot attribute what we see to order as a
cause.

\begin{table}[t]\centering\small
\caption{Both orderings against the arms carrying one repair each, on 8{,}139 ambiguous KoBBQ items and
48 disclosure prompts. All rows are bf16 checkpoints, measured in different sessions across two clusters
with matching engine arguments rather than in one wave; base was re-evaluated on the arm serving
configuration. The KoBBQ denominators differ where an arm drops items: 8{,}139 for base, style, FIN1 and M5, 7{,}991 for safety and 8{,}122 for M4. The conditional share
column is reported for completeness; a between-arm contrast in it is selection-confounded
(\S\ref{sec:selection}). The disclosure column is the strict detector's positive rate pooled over the
shared 48-prompt set (456 rule-pairs per arm, three runs), and the loose column is the second detector's
positive rate on the same pairs. Neither is a compliance estimate (\S\ref{sec:noninvariance}); both are
shown because a single detector cannot carry a cross-arm comparison when detector agreement is itself
arm-dependent. Both detectors are recorded for every arm evaluated on the disclosure axis; the safety
arm was not, and its two disclosure cells are empty rather than imputed.}
\label{tab:order}
\begin{tabular}{llrrrrrr}\toprule
Arm & Stages & Abstention & Abs.\ stereo. & Cond.\ share & Strict & Loose & Dis.\ acc.\\\midrule
base       & ---                        & 81.41\% & 16.02\% & 86.19\% & 0.276 & 0.364 & 88.20\%\\
style      & style                      & 80.10\% & 17.16\% & 86.23\% & 0.061 & 0.140 & 89.82\%\\
safety     & style$\to$safety           & 92.80\% & 6.70\%  & 93.04\% & --- & --- & 88.57\%\\
FIN1       & style$\to$compliance       & 87.82\% & 10.71\% & 87.99\% & 0.945 & 0.952 & 86.24\%\\
M4         & style$\to$safety$\to$compl.& 91.79\% & 7.56\%  & 92.05\% & 0.956 & 0.961 & 88.65\%\\
M5         & style$\to$compl.$\to$safety& 96.98\% & 2.73\%  & 90.24\% & 0.893 & 0.919 & 86.84\%\\\bottomrule
\end{tabular}
\end{table}

Both orderings compose partially, and their observed off-target deviations again lie on emission
variables --- though the M4-versus-safety KoBBQ contrasts are not resolved under template clustering. M4 puts compliance last. On the shared 48-prompt set its
strict-detector disclosure rate is 0.956 against FIN1's 0.945 and its loose rate 0.961 against 0.952.
Clustered on the 40 obligated prompts, neither detector distinguishes the pair --- $+1.10\pp$, CI
$[-1.95, +4.30]$, strict and $+0.88\pp$, CI $[-1.55, +3.61]$, loose --- so both instruments agree that
the disclosure stage survives being applied after safety and neither supports saying it does better. A
random-intercept logistic on the same pairs reaches the opposite verdict on the log-odds scale. The two
target different estimands, on different scales, under different modelling assumptions; we report both,
and the marginal contrast in emission rate is the one the deployment-visible quantity here requires. The cost appears as answering, but not detectably at this
clustering: on the items both arms score, M4 answers $0.68\pp$ more of them than the safety arm and
emits $0.61\pp$ more stereotyped answers in absolute terms, and both template-clustered intervals cover
zero ($[-0.03, +1.41]$ and $[-0.09, +1.34]$). Under the item-level bootstrap these two contrasts were
distinguished; template clustering withdraws them, and the M4-versus-safety difference is a point
estimate we do not claim to have resolved. The conditional share contrast is not distinguished either,
and would be selection-confounded if it were. M5 puts safety last,
and that stage overshoots: it abstains on 96.98\% of ambiguous items against the safety arm's 92.80,
a paired difference of $4.29\pp$ with a template-clustered interval of $[+1.92, +7.24]$ that excludes
zero, and against M4's 91.79; its absolute stereotyped exposure falls to 2.73\%. Both deviations increase
abstention and reduce absolute stereotyped exposure on KoBBQ. What it costs is disclosure, and both detectors register it: on the same
shared set its strict rate is 0.893 and its loose rate 0.919, $5.26\pp$ and $3.29\pp$ below FIN1 and
$6.36\pp$ and $4.17\pp$ below M4. Clustering on the 40 obligated prompts rather than treating the 456
rule-pairs as independent --- the same unit \S\ref{sec:dose} uses --- three of those four contrasts
exclude zero: against FIN1 the strict interval is $[-9.56, -1.16]$ and against M4 the strict and loose
intervals are $[-10.68, -2.38]$ and $[-8.23, -0.66]$. The fourth, M5 against FIN1 under the loose
detector, covers zero at $[-7.14, +0.44]$, so the two detectors agree on the direction but only one of
them resolves that pair.

The two orderings differ by $4.83\pp$ on absolute stereotyped exposure. The right reference for a
DPO-stage difference is the DPO stage's own seed spread, which we measured by retraining the safety
stage under two further seeds: $1.535\pp$ over the 7{,}036 items all three scored (SD $0.792$),
$1.941\pp$ over all 8{,}139 asked, $1.701\pp$ on each arm's own scorable set. The order gap is therefore
$2.5$ to $3.2\times$ the observed DPO seed spread under these denominator conventions, rather than the
$7\times$ that the SFT-stage spread of $0.676\pp$ would suggest. Three seeds give an observed spread,
not a floor, and it is still one run per ordering: the seed replication measures rerunning \emph{one}
order, not the spread of each. The statement this supports is that the two orderings produced clearly different models,
not that stage order is established as the cause.

Two things from that replication belong here because they are \S\ref{sec:rq3}'s argument reappearing
inside a seed experiment. Reseeding moves parse failure by four to five times --- 1.8\% for the shipped
checkpoint against 7.2\% and 9.3\% for the two reseeds --- while re-measuring the shipped checkpoint in
the same session moved it from 148 items to 150. The answered set is a property of the checkpoint rather
than of the harness, so any table over this benchmark has to name its denominator convention
(\S\ref{sec:selection}). And the composition residuals across the three seeds are $0.046$, $0.089$ and
$0.092\pp$: variation under reseeding again falls primarily in the answer-propensity term while the
observed composition residual stays small, the same allocation pattern \S\ref{sec:rq2} reports under
treatment. As there, we do not read the small residual as invariant latent preference, for the reason
\S\ref{sec:selection} gives.

\subsection{A third emission axis: surface form}

One more thing the safety stage did, on prompts with nothing to do with safety. On the 637 held-out
prompts of Table~\ref{tab:style}, non-Korean-script contamination rises to 20.72\% of responses from the
style arm's 4.87\%, and a per-script breakdown attributes the whole increase to simplified Chinese
fragments: 30.74 characters per 10{,}000 against the style arm's 5.08 and base's 2.21, six times its own
starting checkpoint, while Latin contamination over the same arms does not move (13.0\% against 12.2\%
of responses). The compliance stage puts it back: FIN1 reads 5.35 per 10{,}000. We treat this as a
surface-form change; by itself, it does not identify a preference or capability change.

The safety arm is also the only KoBBQ arm that drops items --- 1.82\% of ambiguous items unparseable,
against zero for base and style, falling to 0.21\% after the compliance stage. The two are tempting to
merge and are not one phenomenon: in the 16-token scoring window there are zero Sino-Korean characters in
all 416 excluded completions and zero in a context-matched control. What produces the exclusions is that
this arm answers in prose that breaks the response format, compounded by a scoring regex whose
word-boundary convention cannot see an option letter attached directly to Hangul --- 146 of the 416
excluded completions named a letter the instrument then discarded. The exclusion asymmetry is real, its
cause is surface form, and it is why every contrast involving this arm is classified as overlapping and
tested accordingly.

\section{Limitations}\label{sec:limits}

The bias axis rests on one benchmark family in one language on one base model. We have not shown the
allocation result holds for disambiguated items, other bias benchmarks, or other model families. We
recomputed the same decomposition on English BBQ from stored outputs, and it cannot answer the question:
the base model already abstains on 95.6\% of the ambiguous items, leaving 46 to 68 answered items per
arm and a minimum detectable difference of 25 to $27\pp$ on the conditional share against observed
differences of 3.7 to $5.7\pp$. The English axis neither reproduces nor refutes the Korean result, and
the compliance arm has no stored English data.

Every training arm is $n$=3 or smaller. At three against three the exact permutation test's two-sided $p$
cannot go below 0.100, so no arm comparison in \S\ref{sec:rq1} can carry a small $p$-value by
construction, and the evidence is carried by the sign reversal, the observed seed-range non-overlap and the
template-clustered between-arm intervals instead.
The order comparison in \S\ref{sec:rq4} is $n$=1 per ordering. The two inference axes should not be
confused: the item-level tests quantify uncertainty conditional on each trained checkpoint and do not
substitute for uncertainty across training runs, so any generalisation across training stochasticity
rests only on the three independent seeds per arm.

The lever behind the bias-axis direction is unresolved. Target length and the broader style rewrite
remain entangled, and the fourth arm built to
separate them (\S\ref{sec:rq1}) does not settle it either: its three seeds are too dispersed to
identify a mechanism, and it additionally varies turn-final question structure and the generator's
conditioning distribution alongside hedging. The defensible stopping point is that target form matters
and this study does not identify the individual feature responsible.

The allocation result of \S\ref{sec:rq2} is descriptive of the contrasts we ran and is not a law. Nothing
prevents a treatment from moving the composition term substantially, and a treatment that did would
invalidate no identity, only the summary. The share statistic is also undefined in practice when
$\Delta S$ is near zero. Under template-clustered inference four of the seventeen training contrasts
have intervals covering zero, as does the W4A16 precision contrast; on length-matched seed 2 the ratio
reads $-17$ percent on a composition term of $+0.072\pp$, the two terms having opposite signs on a
$\Delta S$ of $+0.061\pp$, and on W4A16 it reads 126 percent on a composition term of $+0.056\pp$. That is why percentage points are the primary presentation here and the ratio
is quoted only where $\Delta S$ resolves.

The dose-response design shows length is sufficient to produce most of the disclosure effect, not that it
was the causal path; assignment bounds the mediation question from one side only.

Neither detector has been validated against human semantic judgments, so we do not treat
detector-positive rates as calibrated estimates of regulatory compliance. Our claims on this axis are
restricted to detector-measured disclosure emission, intervention effects that reproduce under both
detectors, and cross-arm detector non-invariance.

The precision arms of \S\ref{sec:rq2} are a single run each with unmeasured requantization build spread; a
rebuild floor of $3.56\pp$ measured on a different task elsewhere in this project is why that matters.
They abstain between 89.8\% and 92.9\%, so their magnitude should not be carried to a low-abstention arm, and the
three parsed slightly different item sets (7{,}867 of 8{,}139 in common). We make no claim about quantizer
quality and none about which quantizer to deploy.

The response-style profile of Table~\ref{tab:style} has two known coverage problems, both stated in the
table: base length is right-censored at the generation cap, so length gaps against base are lower bounds;
and the register feature scores different fractions of each arm's responses, so its row is
conditional-on-scored rather than a population share. We report no combined style score, so no reader can
be misled by a weighted average of features with unequal coverage.

$T$=0 is not determinism. Most KoBBQ checkpoint measurements are single evaluations. One checkpoint was
repeated within one session on the same node, moving its absolute stereotyped exposure by $0.013\pp$,
and separately across an H200 and a B200, differing by $0.098\pp$; the disclosure axis is three runs per
arm throughout. Neither repeat estimates variability across training runs, which is why the seed
replications and not the repeats carry the generalisation.

Finally, the training corpus is not released. Weights and evaluation harnesses are; we describe the corpus
construction in enough detail to reproduce the method rather than the file.

\section{Conclusion}\label{sec:conclusion}

Response-style alignment in this pipeline changed two behaviours it never targeted, and in both cases the
deployment-visible changes were carried primarily by how often and how much the model emitted; our
design does not identify whether latent conditional preferences changed. On the bias axis it
answers ambiguous questions more often, and the change in what a user encounters is accounted for
almost entirely by that. On the disclosure axis it says less, and most of the detector-measured loss of
required language is reproduced when response length is assigned directly. Against a base re-evaluated on the same serving configuration, controls trained on
the same prompts and recipe with unmodified teacher responses as targets move the first axis in the
opposite direction from the style arms, the observed seed ranges not overlapping and their means
differing by $2.34\pp$, so what
answer propensity depends on is the form of the training target rather than the prompt set or the recipe
alone --- though which feature of that form is responsible our four-arm design does not identify.

The decomposition that organises the bias result is an identity, and we are explicit about that. What is
empirical is where the movement went: across seventeen training-checkpoint contrasts the
answer-propensity term moves by up to $13.4\pp$ while the composition term never leaves the band from
$+0.00$ to $+0.45\pp$, and two further contrasts produced by changing a quantizer rather than by
training sit in the same place. That is a statement about a deployment-visible quantity, not about
the model's latent preferences, and we stop there deliberately --- the conditional term is computed on
answered subsets the treatment itself selects.

Two things follow for practice. A deployment-relevant bias or disclosure measure can be moved by an
alignment stage that never mentions it, through a variable --- how often and how much the model speaks ---
that a share-based metric does not report. And the instruments that would catch it have to survive the
treatment: a contrast in a ratio conditioned on the answered set is selection-confounded once the treatment
changes that set, and a detector keyed on surface form cannot be assumed to behave invariantly when the
treatment changes surface form. Both are measurable before they are trusted, and in our case
measuring them changed what we were entitled to say.

\appendix

\section{The fourth control arm}\label{app:fourth}

The pre-specified rule for the short-and-hedging arm classified each seed by the nearest of three
landmarks --- \textsc{neutral} at $-1.40$, \textsc{additive} at $-0.10$, \textsc{style} at $+0.94\pp$ ---
and declared the arm unresolved unless the three seeds agreed. They did not: on the base against which
the landmarks were fixed the three seeds classify as \textsc{additive}, \textsc{style} and
\textsc{neutral} respectively. Re-evaluating the base on the arm serving configuration shifts all four
arms by a common constant (the fourth arm's seeds become $-0.639$, $+1.708$ and $-2.519\pp$) and leaves
that classification unchanged.

The rule was mis-specified for an arm this dispersed, and we record that rather than credit it with
firing. With decision bands roughly $1.2\pp$ wide against a per-seed standard deviation of $2.12\pp$, an
arm whose true mean sat exactly on any landmark would still fail seed unanimity most of the time, so the
rule cannot return a verdict here wherever the true mean sits. A rule keyed to the three-seed interval
instead of to per-seed unanimity would have reported the additive band and flagged low power. The
conclusion is unchanged either way, because every pairwise contrast involving this arm overlaps on its
own: $-1.30\pp$ against style at exact permutation $p\approx0.4$, $+1.04\pp$ against neutral at
$p\approx0.6$. The arm mean sits $0.26\pp$ from the additive prediction, which is $0.21$ standard errors
of a three-seed mean whose standard error is $1.22\pp$; we read it as evidence about the additive model
in neither direction.

Two further observations. The dispersion is not instrument variation: across the three seeds the number
of answered items spans 344 of 8{,}139, against 72 for style, 70 for length-matched and 17 for neutral,
and this arm has the highest held-out loss of the four (1.24--1.26 against 0.88--0.90 for style,
0.76--0.78 for length-matched and about 0.74 for neutral). A target that must be short \emph{and} carry
the teacher's reservation is the hardest of the four to fit in one epoch, and if such targets place the
model at a genuinely unstable point --- hedging pulling toward abstention and brevity toward commitment
--- then a two-factor additive account would be mis-specified rather than merely untested. The data
permit that reading and do not establish it, and it too rests on three seeds.

Finally, this arm's mean of $-0.483\pp$ sits $0.004\pp$ from the length-matched arm's $-0.487$ --- a
gap that a common base shift leaves unchanged, so it holds on either base. The two
arms can therefore be read as the same measurement --- both ``not style, not neutral, somewhere in
between'' --- so the fourth cell may have added a second observation of ambiguity rather than resolving
it. Nothing in the design rules that out.

\section{Instrument defects behind body claims}\label{app:instruments}

We audited every instrument supporting a quantitative claim in the main text for provenance and
cross-arm applicability: whether the instrument the surrounding prose describes is the instrument that
produced the number, and whether an instrument carrying a between-arm claim is applied uniformly across
the arms it compares. Where it is not --- the one arm whose parse failures give it a different scored
set --- the affected contrasts are classified as overlapping and tested accordingly (\S\ref{sec:rq4}).
Table~\ref{tab:taxonomy} lists the defects that bear directly on a claim in this paper, with the section
that carries the consequence. Each was found by asking a question the instrument had not been built to
answer, and none raised an error.

\begin{table}[h]\centering\small
\caption{Instrument defects that changed something a reader sees, and where the consequence is carried.}
\label{tab:taxonomy}
\setlength{\tabcolsep}{4pt}
\begin{tabular}{@{}p{3.3cm}p{7.0cm}p{4.6cm}@{}}\toprule
Class & Instance & Consequence in the body\\\midrule
Tautology mistaken for a finding & The threshold/composition split, whose remainder is identically
$A_1\Delta L$ & \S\ref{sec:rq2} reports an allocation, not a fit\\
Instrument substitution & Prose described the five-rule gate; the disclosure figures came from the
looser detector & Both detectors are named and reported (\S\ref{sec:noninvariance})\\
Measurement non-invariance & Agreement between the two disclosure detectors runs 0.44--0.99 by arm &
\S\ref{sec:noninvariance} is built on their joint distribution\\
Protocol provenance mismatch & A flag documented as varying the answer-option layout varies the item
subsample instead & \S\ref{sec:rotation} measures rotation directly\\
Locale-blind convention & A word-boundary regex cannot see an option letter attached to Hangul & The
safety arm's exclusions, \S\ref{sec:rq4}\\
Stack-crossing contrast & A comparison crossing weight precision, KV dtype, speculative decoding and run
date & Precision arms are separated from training contrasts (\S\ref{sec:rq2})\\\bottomrule
\end{tabular}
\end{table}

Two are worth a sentence beyond the table. The flag we described as varying the answer-option layout
takes no part in the layout at all: the rotation is \texttt{md5(sample\_id) \% k}, and the stored layouts
of all 16{,}240 benchmark items reproduce exactly under that function. What the seed argument does is
shuffle the item pool before a per-category cap truncates it, and six of twelve categories sit exactly at
the 2{,}000 cap, holding 12{,}000 of the 16{,}240 scored items, so 73.9\% of the benchmark is a
seed-dependent subsample and two runs under different seeds are not paired. And a verification step meant
to confirm that training hyper-parameters had arrived in the job read back \texttt{adapter\_config.json}
from the output root, while both trainers write that file to \texttt{checkpoint-final/} inside it; no run
could have passed, and completed training runs were logged as failures.

\section*{Reproducibility and provenance}

Weights, model cards and evaluation harnesses are released; the paired training corpus is not. Every number in
this paper resolves to a measurement ledger recording the checkpoint, cluster, serving configuration, item set
and analysis script, and quantities whose ledger does not exist are marked missing there and omitted here rather
than reproduced and inserted. Serving arguments were read back from live pods and the served model identity
asserted before recording for every arm reported in \S\ref{sec:rq1}, \S\ref{sec:rq2} and \S\ref{sec:rq4}.
Statistical tests are computed by one shared module, which selects the admissible test from the stored
item-identifier sets rather than from the design as described; uncertainty over KoBBQ items is
estimated with the KoBBQ template as the resampling or clustering unit, and the recomputation script is
released with the harness. The \texttt{all-test} snapshot we evaluate has SHA-256
\texttt{42efee64\dots f3c00fc3}; with the build unidentified, that hash is the provenance we can offer.

\bibliographystyle{plainnat}
\bibliography{references}

\end{document}